\documentclass{article}

\usepackage{arxiv}

\usepackage[utf8]{inputenc} 
\usepackage[T1]{fontenc}    
\usepackage{hyperref}       
\usepackage{url}            
\usepackage{booktabs}       
\usepackage{amsfonts}       
\usepackage{nicefrac}       
\usepackage{microtype}      
\usepackage{lipsum}

\usepackage{amsmath}
\usepackage[ruled,vlined]{algorithm2e}
\usepackage{graphicx}

\title{Data Proxy Generation for Fast and Efficient Neural Architecture Search}

\author{
  Minje Park \\
  Intel Corporation\\
  \texttt{minje.park@intel.com} \\
}

\begin{document}
\maketitle

\begin{abstract}
Due to the recent advances on Neural Architecture Search (NAS), it gains popularity in designing best networks for specific tasks. Although it shows promising results on many  benchmarks and competitions, NAS still suffers from its demanding computation cost for searching high dimensional architectural design space, and this problem becomes even worse when we want to use a large-scale dataset. If we can make a reliable data proxy for NAS, the efficiency of NAS approaches increase accordingly. Our basic observation for making a data proxy is that each example in a specific dataset has a different impact on NAS process and most of examples are redundant from a relative accuracy ranking perspective, which we should preserve when making a data proxy. We propose a systematic approach to measure the importance of each example from this relative accuracy ranking point of view, and make a reliable data proxy based on the statistics of training and testing examples. Our experiment shows that we can preserve the almost same relative accuracy ranking between all possible network configurations even with 10-20$\times$ smaller data proxy.
\end{abstract}


\section{Introduction}

\begin{figure}
  \centering
  \includegraphics[width=0.6\linewidth]{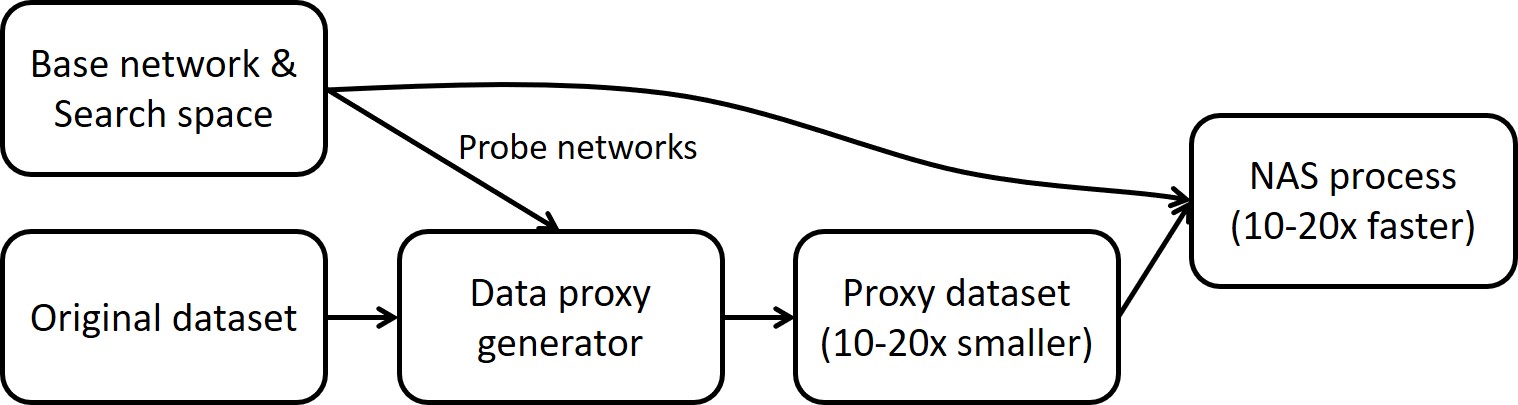}
  \caption{Data proxy generator analyzes the original data based on the given base network and search space to preserve the relative accuracy ranking between possible network configurations with a small set of probe networks. Our experiment with ResNet and CIFAR shows 10-20$\times$ smaller data proxy works for searching the best network.}\label{fig:intro}
\end{figure}

Neural Architecture Search (NAS) is an emerging technique for automating the design of artificial neural networks. Although it shows promising results on diverse benchmarks and competitions, it suffers from its demanding computation cost for searching high dimensional architecture design space. This problem becomes even worse when we want to use large-scale dataset. For example, it takes hours or days to get the best model for CIFAR dataset, which is commonly known as a small dataset compared to other practical datasets (e.g. ImageNet), even with multiple high-end GPU system. In this paper, we propose a systematic approach for creating a data proxy and increasing the efficiency of NAS process by an order of magnitude. Data proxy, if properly constructed, can be applied to any kind of NAS methods. Our basic observation is that each example of training data has a different impact on NAS process from a relative accuracy ranking perspective. That means some subsets of examples are more important when we compare different architectures than others. We introduce the notion of probe networks for efficiently measuring the impact given with the base network, search space, and target task. Our experiment shows that even with 10-20$\times$ smaller data proxy we can keep the relative accuracy ranking of all possible network configurations. Figure~\ref{fig:intro} illustrates how our data proxy generator boosts NAS process. To the best of our knowledge, our proposed method is the first attempt to make data proxy in a systematical manner.

\section{Related Work}

There have been several ideas for accelerating NAS process. The first approach is to use proxy tasks. They search building blocks with partially-trained models with fewer iterations~\cite{tan2019, tan2019e, cai2018p, bak2017} or reuse the result of training with a smaller dataset which is not a subset of target task but different task (e.g. using CIFAR for searching building blocks for ImageNet classification network)~\cite{tan2019} or make a difficulty-sorted data proxy with resampling~\cite{shl2019, shl2019u}. Although there have been no concrete analysis on this bottom-up approach, this paradigm is widely adopted due to its simplicity and efficiency.

The second approach is to use a set of pre-trained models to estimate the accuracy of an arbitrary model during NAS process~\cite{den2017, ist2018, dai2019}. This approach uses fully-trained models but it does not actually trained a model during NAS but makes an accuracy estimator based on a set of already-trained models. If there are enough number of pre-trained models representing the entire search space properly, this would be a good solution for fast (or even interactive) NAS. However, typical NAS dimension is very high, and it requires a large number of pre-trained models to estimate accuracy precisely. That means it also requires a huge computation for NAS. 

Recently many researchers tries to alleviate these problems by adapting gradient-based methods~\cite{liu2018, bichen2019, zoph2019} and one-shot training with the concept of over-parameterized networks~\cite{brock2018}. These methods reduces computational burden by an order of magnitude compared to the classical reinforcement training methods. Note that our data proxy approach is complementary with most of known methods including gradient-based and/or one-shot methods.

\section{Data Proxy Generation}
Our data proxy problem can be formulated as a constrained optimization problem as follows:
\begin{itemize}
\item Objective: Create a data proxy $D_{proxy}$ which is a subset of the original data $D_{original}$, and its size (number of examples) is not greater than $x\%$ of the original data. 
\item Constraint: For every network configuration pair $(N_i,N_j)$ in the search space, $Accuracy(N_i, D_{proxy})$ should be higher than $Accuracy(N_j, D_{proxy})$ if $Accuracy(N_i, D_{original})$ is higher than $Accuracy(N_j, D_{original})$.
\end{itemize}

\begin{figure}
  \centering
  \includegraphics[width=0.6\linewidth]{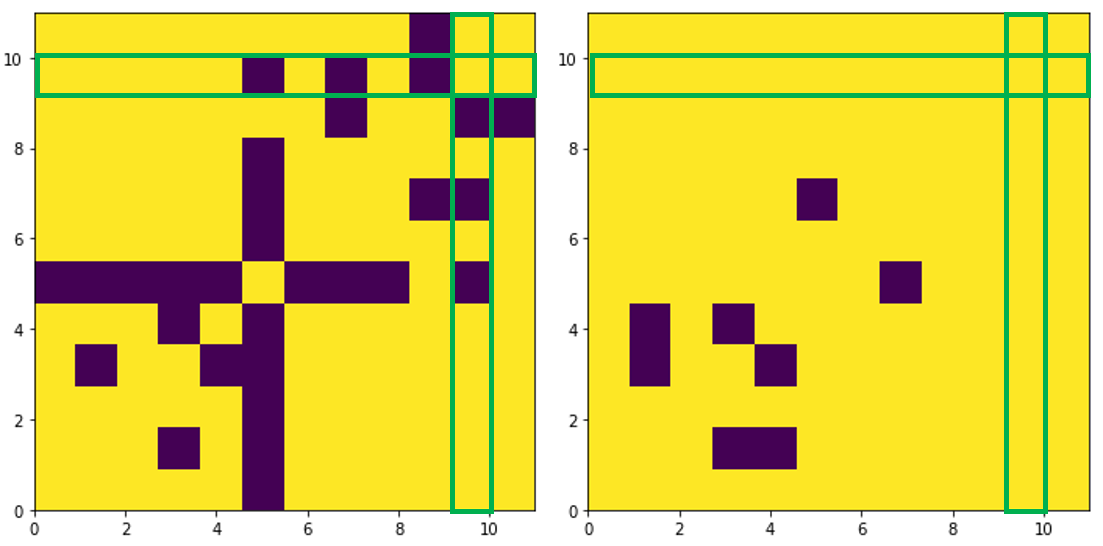}
  \caption{Relative accuracy among 12$\times$12 network configuration pairs of ResNet20 variants on CIFAR10. Yellow cell indicates there is no relative accuracy change, and dark cell indicates relative accuracy with subsampled data has been changed compared to the original training data. (left) Random 5\% subsampling of the original data (right) Data proxy consisting of 5\% of the original data generated by our method. Note that our method maintains the best network configuration (index indicated by green box) while random approach fails.}\label{fig:rel_acc}
\end{figure}

A naïve approach for this problem is to make every possible subset of data and check that the condition holds for every possible network configuration pairs. Obviously this approach is not feasible even for small dataset and search space. Our basic observation on this problem is that some samples are more important than other samples when we rank relative accuracy among possible network configurations within the search space. For example, let's assume that we remove some samples from the training data and draw a complete relative accuracy table between all pairs of network configurations. If removing those samples does not change the relative accuracy table, we can safely discard those samples during NAS process. However, if removing some samples significantly changes the relative accuracy table, we should keep those samples when we make a data proxy. Figure~\ref{fig:rel_acc} illustrates an example of how different subsets affect the relative accuracy ranking.

Based on this observation our next problem is how to effectively measure the importance of each sample from a relative accuracy ranking perspective. Let's assume that we have a complete list of all possible network configurations given with the search space. For each sample $s_{k} \in D_{original}$ we can draw a relative accuracy ranking table $T_{i}$ for this specific sample (actually each cell only contains one of four possible states $\langle {True, False} \rangle$ for a single sample). We can compare this with the original table $T_{ref}$. When we compare these two tables there are four different cases as follows:

\begin{itemize}
\item \textbf{Case1} Samples are correctly classified regardless of network architecture.
\item \textbf{Case2} Samples are incorrectly classified regardless of network architecture.
\item \textbf{Case3} $Accuracy(N_i) > Accuracy(N_j)$ in $T_{ref}$, and sample $s_{k}$ is correctly classified with $N_i$ and incorrectly classified with $N_j$ or vice versa.
\item \textbf{Case4} $Accuracy(N_i) > Accuracy(N_j)$ in $T_{ref}$, but sample $s_{k}$ is incorrectly classified with $N_i$ and correctly classified with $N_j$ or vice versa.
\end{itemize}

Since \textbf{Case1} and \textbf{Case2} provides no information when we measure the importance of each sample, we will focus on \textbf{Case3} and \textbf{Case4}. Our objective is to find a subset of the original data as a data proxy for NAS and we want to keep the ordering of accuracy with this subset. Samples in \textbf{Case3} meet this requirement and thus we will assign higher importance value. Contrarily we need to avoid samples from \textbf{Case4} when we make a data proxy and assign a low importance value for these samples. 

After assigning importance of each sample, we apply stochastic resampling method with the keeping probability of each sample equals to the normalized importance to meet the constraint of the data proxy size as follows:

\begin{center}
$D_{proxy} = Resample(D_{original}, \{ keep\_prob_{i} \})$,
where $keep\_prob_{i} = \frac{Importance_{i}}{\sum_{i \in N} Importance_{i}}$.
\end{center}

Importance of $i$-th sample can be one of four constant values depending on which case this sample belongs to. Basically this can be done by counting the number of configuration pairs. For example we can assign $\{2, 1, 6, 1 \}$ importance values to each case after identifying the case of each sample. This means \textbf{Case3} samples have six times more chances to be resampled in the reduced data proxy. It's not recommended to set the importance values of other cases to zero since we need some fractions of the original data to make the training process converge. Table~\ref{tb:procedure} describes the detailed procedure of our data proxy generation scheme (including probe networks explained in Section~\ref{sec:probe}).

\subsection{Probe Networks}\label{sec:probe}
One remaining problem is that we need to prepare and train all possible network configurations to measure sample importance collectively. To reduce this computation cost, we use a set of $k$ selected networks as representatives, which are called probe networks, rather than using a complete list of all possible network configurations. One of simplest and most natural choice of probe networks is to use two boundary network configurations from a compute complexity perspective. The upper boundary model is the most complex model we can make during search process, and the lower boundary model is the lightest one. There is an obvious trade-off between the number of probe networks and the reliability of measured importance. Throughout our experiment we found that these two boundary probe networks ($k==2$) serve our purpose well. Depending on the dimensionality of search space and the complexity of base network more probe networks can be used if you need more reliability by sacrificing data proxy generation time.

\subsection{Similarity of Examples}
Our goal is to keep the relative ranking of accuracy, and the accuracy is measured with test data. So, we need to measure the importance value not with training examples but with test ones. However, resampling should be done for training data. To set importance values of training examples, we need a similarity mapping function from test examples to training ones. A common choice for defining similarity is using the intermediate layer output of deep neural network and apply standard distance metric functions like Euclidean or cosine distances. To efficiently find the nearest test samples of each training example, we can create a nearest neighbor mapping function based on the features extracted from the upper boundary network (which is the most capable network presumably). Depending on the dimensionality of feature it's recommended to apply dimensionality reduction method such as  Principal Component Analysis before constructing a nearest neighbor mapping function.

\subsection{Reducing the Number of Class Labels}\label{sec:labels}
For some data sets like CIFAR100 or ImageNet, some labels contain too small number of examples. Further reduction on these labels make the training not converged since each output neuron does not generate enough amount of gradient for back propagation. In this case, we can remove the label itself. Just like we set the importance of each train sample, we can use aggregated importance value of each label as the importance of that label, and apply the same logic to reduce the number of class labels. Another possible option is to merge some labels according to its similarity and make a super-label which contains more examples for each label. We use the first approach since merging classes requires robust label-to-label distance metric. For example, if our target data proxy size is 10\% of the original data, we first remove some labels to create a 20\% of the original size with label-wise importance values, and then remove some samples with sample-wise importance values to create the final 10\% data proxy.

\begin{algorithm}[t]
Input: $D_{original}$, $N_{base}$, $SearchSpace$, $ratio$\;
Output: $D_{proxy}$\;

\begin{enumerate}
\item Train a pair of probe networks with the original training data $D_{original}$
\item Probe networks consists of two boundary network configurations of the search space $N_{upper}$ and $N_{lower}$
 
\item For each sample of test (or validation) data $s_i$, set the importance value $Importance^{test}_i$ according to the following criteria:
\begin{itemize}
\item $c1$ when both $N_{lower}$ and $N_{upper}$ correctly classify
\item $c2$ when both $N_{lower}$ and $N_{upper}$ incorrectly classify
\item $c3$ when both $Accuracy(N_{lower}, D_{original})$ is lower than $Accuracy(N_{upper}, D_{original})$ and only $N_{upper}$ correctly classifies
\item $c4$ when both $Accuracy(N_{lower}, D_{original})$ is not lower than $Accuracy(N_{upper}, D_{original})$ and only $N_{lower}$ correctly classifies
\end{itemize}

\item Extract features (intermediate layer output) $F_{i}$ from $\{s_{i}\}$ by using $N_{upper}$
\item Reduce the dimensionality of $F_{i}$ by using PCA (or any decomposition method)
\item Construct a nearest neighbor map $NearestNeighbor_{F}(\cdot)$
\item For each train sample $t_i$, set $Importance^{train}_{i}$ as $Importance^{test}_{NearestNeighbor_{F}(t_{i})}$

\item Set keep probability $keep\_prob_{i}$ of each train sample as $\frac{Importance^{train}_{i}}{\sum_{i \in N} Importance^{train}_{i}}$

\item Resample the original data $D_{original}$ to create a data proxy $D_{proxy}$ with $\{keep\_prob_{i}\}$
	
\end{enumerate}
	
\caption{Data Proxy Generation}\label{tb:procedure}
\end{algorithm}

\section{Experiment}

Our first experiment is how this data proxy affects typical neural architecture search. We used ResNet20 as a base network, and defined the following 12 different network configurations by changing the number of base filters (B), kernel size (K), and multiplier (M)~\cite{he2016}. Table~\ref{tb:r20c10} summarizes the accuracy of each configuration with and without data proxy on CIFAR10~\cite{kriz2009}, which consists of 50000 training samples and 10 classes (each class has 5000 samples). As a baseline for comparison, we used random uniform resampling method. We applied the same optimizer (Adam), batch size, learning rate policy and other training parameters to minimize the impact of different training-related hyperparameters.

When we reduce the training data to 10\%, both approaches worked quite well. The correlation score between the accuracy trained with the original data and that with the proxy data is above 0.97, which is high enough for neural architecture comparison. However, when we reduced the data proxy size to 5\% our data proxy approach works better than random approach. The correlation score is increased from 0.922 to 0.965, which is similar to 10\% data proxy case. More importantly, the best network configuration is consistent with the full data case when we used our data proxy approach but random selection approach fails to find the best configuration (bold text indicates the highest accuracy for each data sampling). When we use 10\% data proxy, training time is reduced from 25sec per epoch to 3sec per epoch which is 8 times faster training speed. The speed gain can further increase when we use 5\% data proxy to 1.6sec (single nVidia GTX1080). Please note that the absolute accuracy is not relevant since our goal is to get the relatively best network architecture among possible network configurations.

\begin{table}[t]
\begin{center}
\begin{tabular}{|l|r|r|r|r|r|r|}
 \hline
 ID & B, K, M &  100\% (original) & 10\% (random) & 10\% (ours) & 5\% (random) & 5\% (ours) \\ \hline
 1 & 8, 3, 2      & 0.832             & 0.5865            & 0.6076 & 0.5343 & 0.4762 \\ \hline
 2 & 8, 3, 4      & 0.8802            & 0.6499            & 0.6552 & 0.5613 & 0.5353 \\ \hline
 3 & 8, 5, 2      & 0.8494            & 0.6197            & 0.6068 & 0.5484 & 0.4922 \\ \hline
 4 & 8, 5, 4      & 0.8792            & 0.6502            & 0.6524 & 0.5803 & 0.5464 \\ \hline
 5 & 16, 3, 2     & 0.8814            & 0.6668            & 0.6788 & 0.5791 & 0.5321 \\ \hline
 6 & 16, 3, 4     & 0.9096            & 0.7147            & 0.7152 & 0.6203 & 0.5821 \\ \hline
 7 & 16, 5, 2     & 0.8873            & 0.6746            & 0.6823 & 0.5912 & 0.5551 \\ \hline
 8 & 16, 5, 4     & 0.9069            & 0.7044            & 0.7161 & 0.6372 & 0.595 \\ \hline
 9 & 32, 3, 2     & 0.9008            & 0.6983            & 0.6998 & 0.5967 & 0.5613 \\ \hline
 10 & 32, 3, 4    & \textbf{0.9123}   & \textbf{0.7358}   & \textbf{0.7302} & 0.6309 & \textbf{0.6161} \\ \hline
 11 & 32, 5, 2    & 0.9049            & 0.716             & 0.7048 & 0.6401 & 0.5782 \\ \hline
 12 & 32, 5, 4    & 0.9096            & 0.7305            & 0.7295 & \textbf{0.6428} & 0.5964 \\ \hline
 \multicolumn{2}{|c|}{Correlation score} & 1.0 & 0.976 & 0.972 & 0.922 & 0.965 \\ \hline
\end{tabular}\caption{ResNet20 variations on CIFAR10 with and without data proxy}\label{tb:r20c10}
\end{center}
\end{table}

In our next experiment, we used CIFAR100~\cite{kriz2009} which consists of 100 classes and each class contains 500 samples. If we reduce each class independently as we did in CIFAR10, there remains average 50 samples per class and training is not converged since each output neuron can not generate enough amount of gradient for back propagation. As we explained in Section~\ref{sec:labels}, we first reduced the number of samples to 50\%, and then dropped the labels which have lower keep probability samples to meet the final 10\% target (20 classes are left). We used the same base network and configurations with our first experiment (refer to Table~\ref{tb:r20c10}). As shown in Table~\ref{tb:r20c100}, both approaches work quite well but our data proxy scheme is still better than random selection and finds the same best network configuration of full data training. Unfortunately, when we reduced the data to 5\% of the original size both approaches were not converged.

\begin{table}[t]
\begin{center}
\begin{tabular}{|l|r|r|r|r|}
 \hline
 ID & B, K, M &  100\% (original) & 10\% (random) & 10\% (ours) \\ \hline
 1 & 8, 3, 2      & 0.4971   & 0.6128 & 0.55 \\ \hline
 2 & 8, 3, 4      & 0.6095   & 0.667  & 0.6343 \\ \hline
 3 & 8, 5, 2      & 0.5319   & 0.6454 & 0.6107 \\ \hline
 4 & 8, 5, 4      & 0.6097   & 0.6691 & 0.656 \\ \hline
 5 & 16, 3, 2     & 0.6244   & 0.6844 & 0.6768 \\ \hline
 6 & 16, 3, 4     & 0.6679   & 0.705  & 0.7098 \\ \hline
 7 & 16, 5, 2     & 0.6324   & 0.6713 & 0.6777 \\ \hline
 8 & 16, 5, 4     & 0.6644   & 0.726  & 0.7002 \\ \hline
 9 & 32, 3, 2     & 0.6609   & 0.6987 & 0.6996 \\ \hline
 10 & 32, 3, 4    & \textbf{0.6959}   & 0.7249 & \textbf{0.756} \\ \hline
 11 & 32, 5, 2    & 0.6533   & 0.7116 & 0.7073 \\ \hline
 12 & 32, 5, 4    & 0.6907   & \textbf{0.7327} & 0.7409 \\ \hline
 \multicolumn{2}{|c|}{Correlation score} & 1.0 & 0.955 & 0.975 \\ \hline
\end{tabular}\caption{ResNet20 variations on CIFAR100 with and without data proxy}\label{tb:r20c100}
\end{center}
\end{table}

As our final experiment we used a different base network, EfficientNet-B0~\cite{tan2019e}, which is generated by NAS with reinforcement learning. It consists of a set of mobile inverted bottleneck blocks~\cite{tan2019}, and we defined three configuration  hyperparameters for this network, which are subsets of the original search parameters.
\begin{itemize}
\item Repeats: \{[1, 2, 2, 2, 2, 3, 1], [1, 2, 2, 3, 3, 4, 1]\} (two configurations, r1, r2)
\item Kernel sizes: \{[3, 3, 3, 3, 3, 3, 3], [3, 3, 5, 3, 5, 5, 3]\} (two configurations, k1, k2)
\item Number of filters: \{[18, 16, 16, 24, 48, 68, 116], [24, 16, 20, 32, 64, 90, 152], [32, 16, 24, 40, 80, 112, 192]\} (three configurations, f1, f2, f3)
\end{itemize}

We created 12 different network configurations based on this configuration hyperparameters. Table~\ref{tb:ef0c10} summarizes the result of accuracy on CIFAR10 with and without data proxy. Unlike ResNet20 case, we found that the correlation scores are low. We think EfficientNet-B0 has a very deep architecture and CIFAR10 from scratch training does not work well for this base network. However, our data proxy scheme still works much better than random selection, and successfully select the best network configuration consistent with that of using the full training data. The significant difference between the baseline and ours in this challenging setting implies our data proxy generation scheme reflects sample importance properly from a relative accuracy ranking point of view.

\begin{table}[t]
\begin{center}
\begin{tabular}{|l|r|r|r|r|}
 \hline
 ID & Configuration &  100\% (original) & 10\% (random) & 10\% (ours) \\ \hline
 1  & r1, k1, f1    & 0.8767   & 0.6985 & 0.6565 \\ \hline
 2  & r1, k1, f2    & 0.8838   & 0.684  & 0.6629 \\ \hline
 3  & r1, k1, f3    & 0.8934   & 0.7091 & 0.6943 \\ \hline
 4  & r1, k2, f1    & 0.8735   & 0.6862 & 0.6789 \\ \hline
 5  & r1, k2, f2    & 0.885    & 0.728  & 0.655 \\ \hline
 6  & r1, k2, f3    & 0.8917   & \textbf{0.7309} & 0.688 \\ \hline
 7  & r2, k1, f1    & 0.868    & 0.7024 & 0.6861 \\ \hline
 8  & r2, k1, f2    & 0.8746   & 0.706  & 0.6442 \\ \hline
 9  & r2, k1, f3    & 0.8905   & 0.6811 & 0.6842 \\ \hline
 10 & r2, k2, f1    & 0.8745   & 0.67   & 0.6688 \\ \hline
 11 & r2, k2, f2    & 0.8913   & 0.6867 & 0.703 \\ \hline
 12 & r2, k2, f3    & \textbf{0.897}    & 0.69   & \textbf{0.7228} \\ \hline
 \multicolumn{2}{|c|}{Correlation score} & 1.0 & 0.171 & 0.612 \\ \hline
\end{tabular}\caption{EfficientNet-B0 variations on CIFAR10 with and without data proxy}\label{tb:ef0c10}
\end{center}
\end{table}

\section{Conclusion}
We proposed a novel method for generating data proxy which can be generally applicable for any kind of neural architecture search. Our method relies on the observation that the importance of each training sample given with search space and configuration differs from each other, and we estimate the importance with a small set of probe networks. Based on the estimated importance, we successfully reduce the size of training data more than 10 times while keeping the relative accuracy almost intact with all possible network configurations. To the best of our knowledge, our work is the first systematic approach for reducing training data for neural architecture search, and we expect to open a new research direction for pursuing the efficiency of NAS from the data optimization point of view.

\bibliographystyle{unsrt}  
\bibliography{references}  

\end{document}